\documentclass[runningheads]{llncs}

\usepackage{cite}
\usepackage{amsmath,amssymb,amsfonts}
\usepackage{algorithmic}
\usepackage{graphicx}
\usepackage{textcomp}
\usepackage{xcolor}
\usepackage{listings}
\usepackage{todonotes}
\usepackage{hyperref}
\usepackage{subcaption}
\usepackage{tikz}
\usepackage{wrapfig}

\begin{document}

\title{GERD:\\Geometric event response data generation}


\author{Jens Egholm Pedersen$^*$\inst{1,2}\orcidID{0000-0001-6012-7415} \and
Dimitrios Korakovounis$^*$\inst{2}\orcidID{0009-0006-2490-3206} \and
Jörg Conradt\inst{2}\orcidID{0000-0001-5998-9640}
}
\authorrunning{J. E. Pedersen et al.}

\institute{DTU Electro, Technical University of Denmark, Copenhagen, Denmark \email{jegpe@dtu.dk} \and
Department of Computational Science and Technology, KTH Royal Institute of Technology, Stockholm, Sweden \email{dimkor@kth.se}, \email{conr@kth.se}
\\
$^*$ shared first-authorship
}


\maketitle

\begin{abstract}
Event-based vision sensors offer high temporal resolution, high dynamic range, and low power consumption, yet event-based vision models lag behind conventional frame-based vision methods.
We argue that this gap is partly due to the lack of principled study of the transformation groups that govern event-based visual streams.
Motivated by the role that geometric and group-theoretic methods have played in advancing computer vision, we present GERD: a simulator for generating event-based recordings of objects under precisely controlled affine, Galilean, and temporal scaling transformations.
By providing ground-truth transformations at each timestep, GERD enables hypothesis-driven and controlled studies of geometric properties that are otherwise hard to isolate in real-world datasets or with current event simulators.
GERD supports three noise models and sub-pixel motion as a complement to real sensor datasets.
We demonstrate its use in training by evaluating models from the literature with geometric guarantees and release GERD as an open tool available at \href{https://github.com/ncskth/gerd}{github.com/ncskth/gerd}.
\end{abstract}


\section{Introduction}
Event-based vision sensors capture sparse, asynchronous data, providing significant advantages in temporal resolution, dynamic range, and power efficiency.
But the sparse and discrete data from event sensors cannot immediately benefit from the numerous and impressive computer vision results relying on dense and real-valued Red-Green-Blue (RGB) frames.
The goal of a machine learning model is to generalize from its training data to unseen data from the same setting and an important factor in model generalization is an innate respect for the structures in the data---for instance, whether an object is considered to be the same wherever it appears (translation invariance).
Building this into models in the form of geometric priors dramatically improves their performance \cite{bronstein2021geometric}, demonstrated for translation \cite{Lecun_Bottou_Bengio_Haffner_1998} and broader symmetry groups \cite{Cohen_Welling_2016}.
To advance event-based models, this paper argues that we must first understand their geometric priors from first principles, with a focus on the symmetries of spatio-temporal ``point clouds'' rather than the symmetries of dense frames.
Building that understanding is an empirical task requiring datasets and metrics that isolate individual symmetries, so a model's sensitivity to each can be directly measured.

We contribute a simulator that generates event-based recordings of objects subject to carefully controlled transformations.
Crucially, GERD provides the ground-truth transformation parameters at every timestep, giving full access to the affine transformation matrix $T$ for each shape throughout the recording (Figure \ref{fig:pipeline}).
This enables quantitative evaluation of geometric properties that would be impossible to isolate in real-world recordings.
The resulting data is well-suited to test the robustness and generalization of event-based computer vision models.
We designed our simulator to strongly resemble data from an event-camera, but the two should not be considered equivalent.
Instead, we focus on studying the spatio-temporal structure of event data streams under controlled conditions as a prerequisite for the processing of real-world data.
Specifically, we restrict our scope to 2D temporal geometric transformations; optical effects such as lens distortion, motion blur, and other camera-specific artifacts are not modeled, nor do we consider 3D scene geometry.
These simplifications are deliberate: they isolate the geometric structure of event streams from sensor-specific confounds, providing a clean testbed for studying transformation properties.
\begin{figure}
    \vspace{-3em}
    \centering
\usetikzlibrary{arrows.meta,calc,backgrounds}

\definecolor{gerdcol}{HTML}{0F6E56}
\definecolor{scenecol}{HTML}{6F0F56}

\begin{tikzpicture}[
  font=\sffamily,
  box/.style={rounded corners=3pt, draw, line width=0.5pt,
              minimum width=2.9cm, minimum height=1.05cm, align=center, inner sep=2pt},
  gerd/.style={box, draw=gerdcol, fill=gerdcol!9},
  scene/.style={box, draw=scenecol, fill=scenecol!10},
  shared/.style={box, draw=black!60, fill=black!4},
  flow/.style={-{Stealth[length=2mm,width=1.6mm]}, line width=0.6pt, draw=black!60},
  sheet/.style={rounded corners=3pt, draw=gerdcol, line width=0.5pt},
]

\begin{scope}[on background layer]
  \draw[sheet, draw opacity=0.30] ($(1.0,1.475)+(0.24,0.26)$) rectangle ($(3.9,2.525)+(0.24,0.26)$);
  \draw[sheet, draw opacity=0.50] ($(1.0,1.475)+(0.12,0.13)$) rectangle ($(3.9,2.525)+(0.12,0.13)$);
  \draw[sheet, draw opacity=0.30] ($(4.5,1.475)+(0.24,0.26)$) rectangle ($(7.45,2.525)+(0.24,0.26)$);
  \draw[sheet, draw opacity=0.50] ($(4.5,1.475)+(0.12,0.13)$) rectangle ($(7.45,2.525)+(0.12,0.13)$);
\end{scope}

\node[gerd] (g1) at (2.45,2)
  {specify $T(t)$\\[1pt]{\scriptsize\color{black!55}sweep transformations}};
\node[gerd] (g2) at (6,2)
  {render 2D shape\\[1pt]{\scriptsize\color{black!55}transform and rasterise}};

\node[scene] (e1) at (2.45,0)
  {specify scene\\[1pt]{\scriptsize\color{black!55}3D geometry}};
\node[scene] (e2) at (6,0)
  {move camera\\[1pt]{\scriptsize\color{black!55}project 3D $\to$ 2D}};

\node[shared, minimum height=1.7cm, minimum width=2.4cm] (sh) at (9.3,1)
  {event frames\\[2pt]{\scriptsize\color{black!55}via differencing}};

\node[anchor=east, text=gerdcol,  font=\small\bfseries, align=right] at (0.75,2) {GERD};
\node[anchor=east, text=scenecol, font=\small\bfseries, align=right] at (0.75,0) {Scene-\\based};

\foreach \k [evaluate=\k as \op using {1-0.30*\k}] in {0,1,2}{
  \draw[flow, draw opacity=\op] ($(g1.east)+(0.12*\k,0.13*\k)$) -- ($(g2.west)+(0,0.13*\k)$);
}
\foreach \k [evaluate=\k as \op using {1-0.30*\k}] in {0,1,2}{
  \draw[flow, draw opacity=\op] ($(g2.east)+(0.12*\k,0.13*\k)$) -- ($(sh.west)+(0,0.20+0.25*\k)$);
}
\draw[flow] (e1.east) -- (e2.west);
\draw[flow] (e2.east) -- ($(sh.west)+(0,-0.40)$);

\node[anchor=south, text=gerdcol, font=\footnotesize, align=center, text width=7cm]
     at (4.05,2.75) {ground-truth $T(t)$ at every step};
\node[anchor=north, text=scenecol, font=\footnotesize, align=center]
     at (4.05,1.05) {$T(t)$ entangled with scene};
\node[anchor=north, font=\scriptsize, text=black!55] at ($(sh.south)+(0,-0.08)$)
     {shared sensor model};

\end{tikzpicture}
    \caption{Two routes to event data, contrasted by where the geometric transformation enters.
    \textbf{GERD (top)} specifies a transformation $T(t)$ that is parameterized to produce multiple instantiations -- shown by the stacked outlines -- yielding a labelled family of recordings. \textbf{Scene-based simulators (bottom)} specify a 3D scene and move a camera; the image-plane transformation arises only implicitly and is entangled with scene geometry. Both routes terminate in the same intensity-differencing event model, so the methods only differ in the provenance of the motion.}
    \label{fig:event-frame-methods}
    \vspace{-3em}
\end{figure}

\subsection{Related work}

A recent survey on event-based datasets revealed that 423 datasets exist so far \cite{cohen2026land}. 
That number is growing every year, but event datasets are still vastly outnumbered by frame-based computer vision datasets.
The existing event-based datasets can be divided into two classes: (1) recorded using physical sensors or (2) generated in simulation.
For a full overview, we refer the reader to \cite{cohen2026land}.

\subsubsection{Recorded event-based datasets}

The N-MNIST dataset \cite{10.3389/fnins.2015.00437} is an early attempt to transfer the existing MNIST handwritten digits dataset \cite{LeCun2005TheMD} from classical machine learning to event-based vision. Other datasets of continuous signals recorded using event sensors have been developed, such as the DVS-Gesture \cite{8100264}, and SHD and SSC \cite{9311226} that consist of classification tasks for visual and auditory event-based signals respectively. 
Other datasets only contain event streams with annotations \cite{sironi2018hatshistogramsaveragedtime, perot2020learningdetectobjects1} or various modalities such as recording from frame-based vision sensors \cite{gehrig2021dsecstereoeventcamera}, IMU and other sensors \cite{hu2020ddd20endtoendeventcamera}, but no information about the geometrical structure and transformation of the objects has been extracted and provided.

\subsubsection{Event-based generators}
\begin{figure}
    \vspace{-4em}
    \centering
    \includegraphics[width=0.9\linewidth]{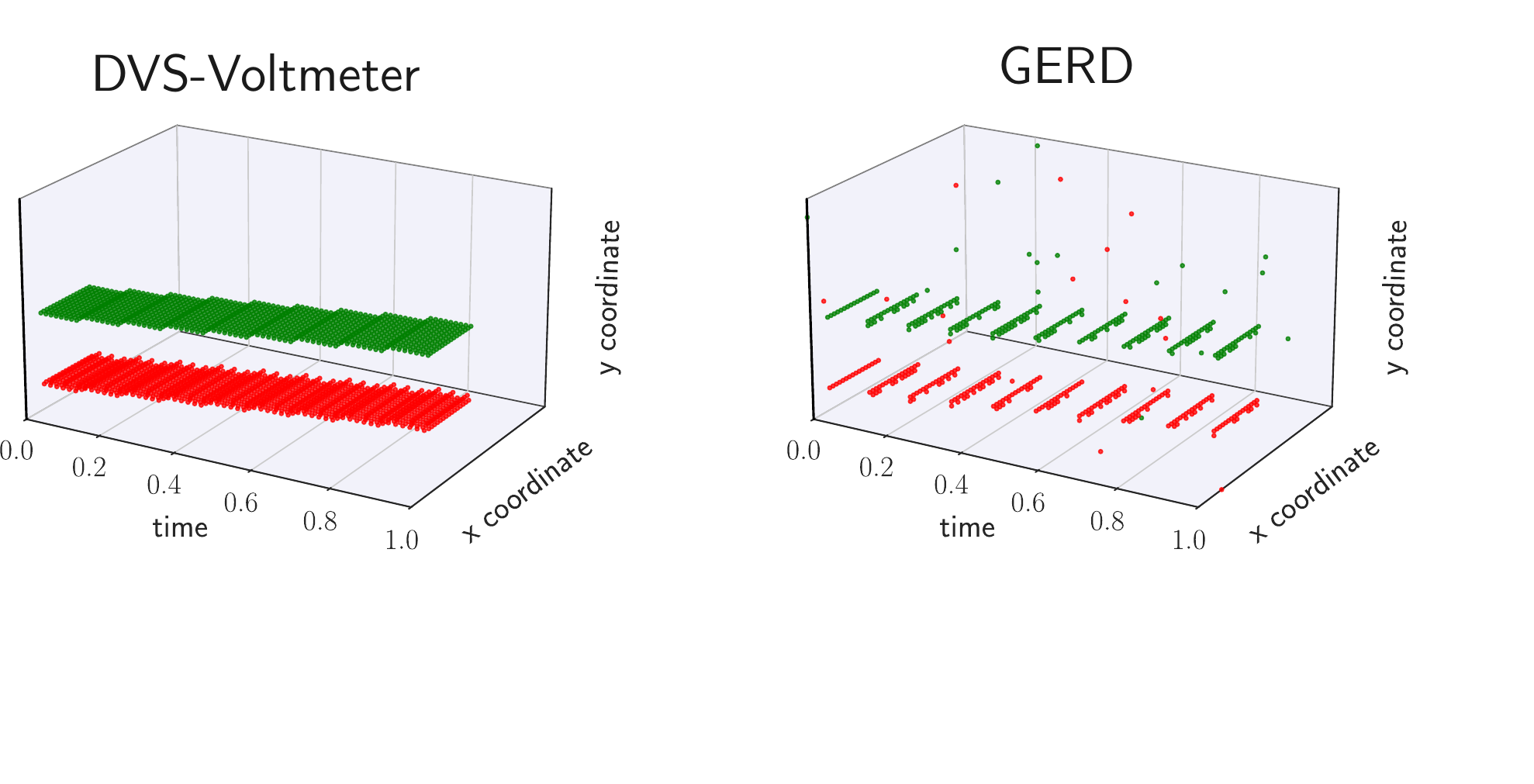}
    \vspace{-4em}
    \caption{Visual comparison between DVS-Voltmeter \cite{DVS-Voltmeter} and our method. The event stream is generated using a square moving across the field of view by 1 pixel per frame (5 ms per frame for DVS-Voltmeter). DVS-Voltmeter produces events at sub-frame time resolution, resulting in a semi-continuous event representation. However, in simple simulation settings, no noise is introduced in the generated events, while our simulator (right) produces discrete-time noisy events.}
    \label{fig:event-voxel}
    \vspace{-1em}
\end{figure}
A different method to obtain event-based data is to artificially generate them. Approaches mostly differ on whether the generation comes as a transformation of existing frame sequences or as generations of new scenes.

Converting sequences of frames to events is usually done by subtracting consecutive frames and applying a threshold. If the difference exceeds the threshold, an event is generated \cite{7862386}. In \cite{8296630}, Bi et al. also account for the contrast around the pixel. However, these approaches do not account for the timing between frames. To address this, Mueggler et al. \cite{Mueggler2016} propose interpolating linearly between the frames and obtaining the timing of the event at the exact time that the log luminance surpasses the threshold. Other methods, such as using rate encoding on the threshold, have been proposed as well \cite{8296630, 7850249}. To produce more realistic event streams, \cite{DVS-Voltmeter} and \cite{hu2021v2evideoframesrealistic} account for characteristics of the event sensors, such as leak currents and their effect on event generation and hot pixel generation, shown in Figure \ref{fig:event-voxel}.
Another family of works uses rendered images to produce realistic event data \cite{6272143, Mueggler2016, pmlr-v87-rebecq18a, Li2018}.
This permits controlling the frame rate, to optionally provide very high frame rates \cite{Li2018}, or providing dynamical sampling in cases where the luminance changes fast or significant pixel displacement is noted \cite{pmlr-v87-rebecq18a}.
In \cite{gehrig2020videoeventsrecyclingvideo}, Gehrig et al. upsample frames in the temporal domain before discretizing the frames to generate events from an event camera simulator (ESIM) \cite{pmlr-v87-rebecq18a}.
Rather than operating on the geometries in the scene, ESIM moves the camera in a 3D space, which is also the approach taken by IEBCS \cite{joubert2021event}.
Transformations in GERD are known at each step (see Figure \ref{fig:pipeline}) whereas scene-based methods such as  \cite{6272143, Mueggler2016, gehrig2020videoeventsrecyclingvideo,pmlr-v87-rebecq18a, joubert2021event} need to decouple it from the scene by applying camera motion and re-projecting onto the scene geometries.
Figure \ref{fig:event-frame-methods} contrasts the two methods.

Deep learning approaches have also been proposed, such as in \cite{zhang2024v2cevideocontinuousevents}, where a UNet is trained to predict the event streams from image sequences and in \cite{zhu2019eventganleveraginglargescale}, where a generative adversarial network is trained to produce realistic event data.
\vspace{-0.5em}
\section{Event generation method}
\begin{figure}
    \vspace{-2em}
    \centering
    \includegraphics[width=\linewidth]{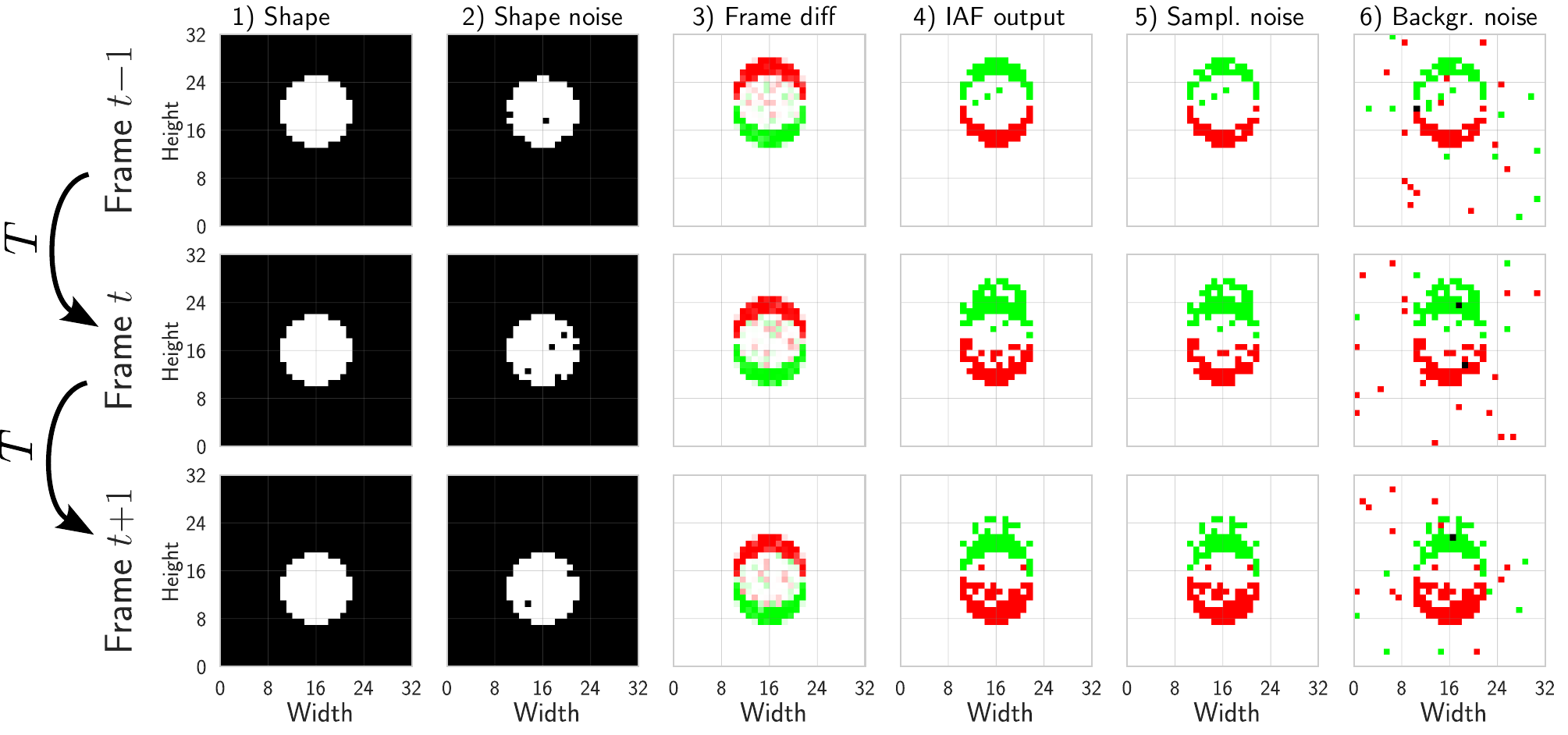}
    \caption{The event generation pipeline for three consecutive frames. 
    First, 1) a circle shape is subject to some transformation $T$, rendered, and 2) sampled with a density parameter.
    Then, 3) the difference between the previous frame is 4) fed to an integrate-and-fire neuron \eqref{eq:fire} to emulate the event-triggering in camera circuits.
    Finally we add 5) sampling and 6) background noise.
    This example uses 5\% noise for the shape, sampling, and background noise and $T$ is a constant vertical translation of 3 pixels.
    }
    \label{fig:pipeline}
    \vspace{-2em}
\end{figure}

We generate events by (1) rendering a shape subject to a given transformation, and (2) taking the difference between the two frames (see Figure \ref{fig:pipeline}).
Basing the simulation around the shape transformation $T$ enables the careful control of each shape across time (see Figure \ref{fig:pipeline}).
In GERD, $T$ can describe any possible affine motion in space, its velocity $T'$, and acceleration $T''$.
Additionally, noise can be added to both the underlying shape, the difference sampling stage, or as general background noise.
\begin{figure}
    \vspace{-0.5em}
    \centering
    \includegraphics[width=.7\linewidth]{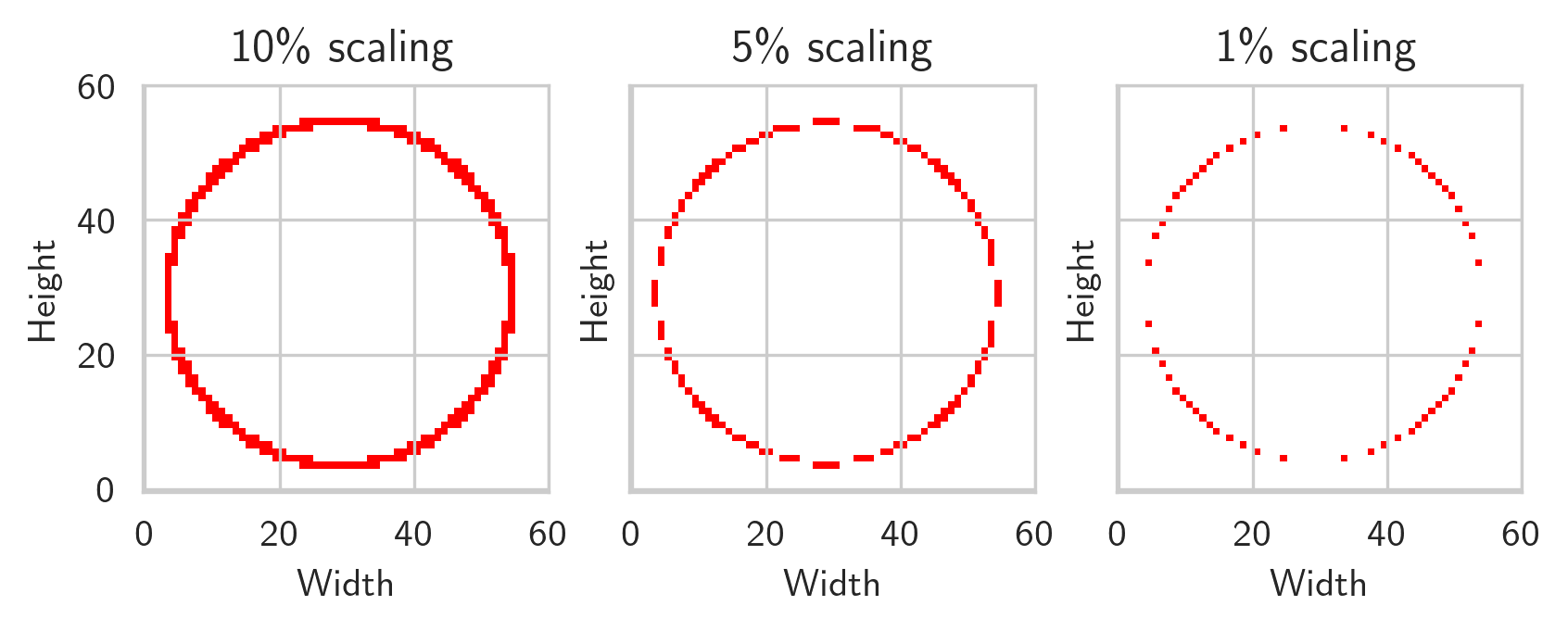}
    \caption{A circle scaled down by 10\%, 5\%, and 1\%.
    Specifying the scaling factor lets us control the amount of emitted events.}
    \label{fig:density}
    \vspace{-1em}
\end{figure}

Most event-camera simulators output asynchronous event tuples $(x, y, t, p)$, representing individual pixel firings with continuous timestamps.
GERD outputs synchronous sparse frames---boolean tensors of shape $(T, 2, H, W)$ denoting the time, polarity, height, and width---where each frame captures all events at a given timestep.
This frame-synchronous format integrates directly with batch-based deep learning workflows without requiring binning or accumulation as a pre-processing step.
Since the sparse representation retains the coordinates of active pixels, asynchronous event tuples can be easily extracted, if needed.
By repeating the procedure above frame-by-frame, we can generate arbitrarily long videos.

Changing the magnitude of the transformations produces arbitrarily sparse recordings, since our transformations determine the amount of change per frame (see Figure \ref{fig:density}).
In real life, subjecting an object to tiny transformations over time produces highly sparse frames, which means that the object is either moving very slowly or that the timesteps between the frames are minimal.
Since the generated frames are not bound by physical time, these interpretations are equivalent, making the dataset suitable for settings with arbitrarily small timesteps or transformational speeds.

The simulator is built using PyTorch \cite{Paszke_2019} and can be accelerated with hardware accelerators supported by PyTorch, such as graphics or tensor processing units.
We represent the frames as sparse tensors using an address-event representation, which can easily be converted to dense frames if needed.
\vspace{-0.9em}
\subsection{Upsampling and integration for sub-pixel motion}
\begin{wrapfigure}{r}{0.6\textwidth}
    \vspace{-2.5em}
    \centering
    \includegraphics[width=\linewidth]{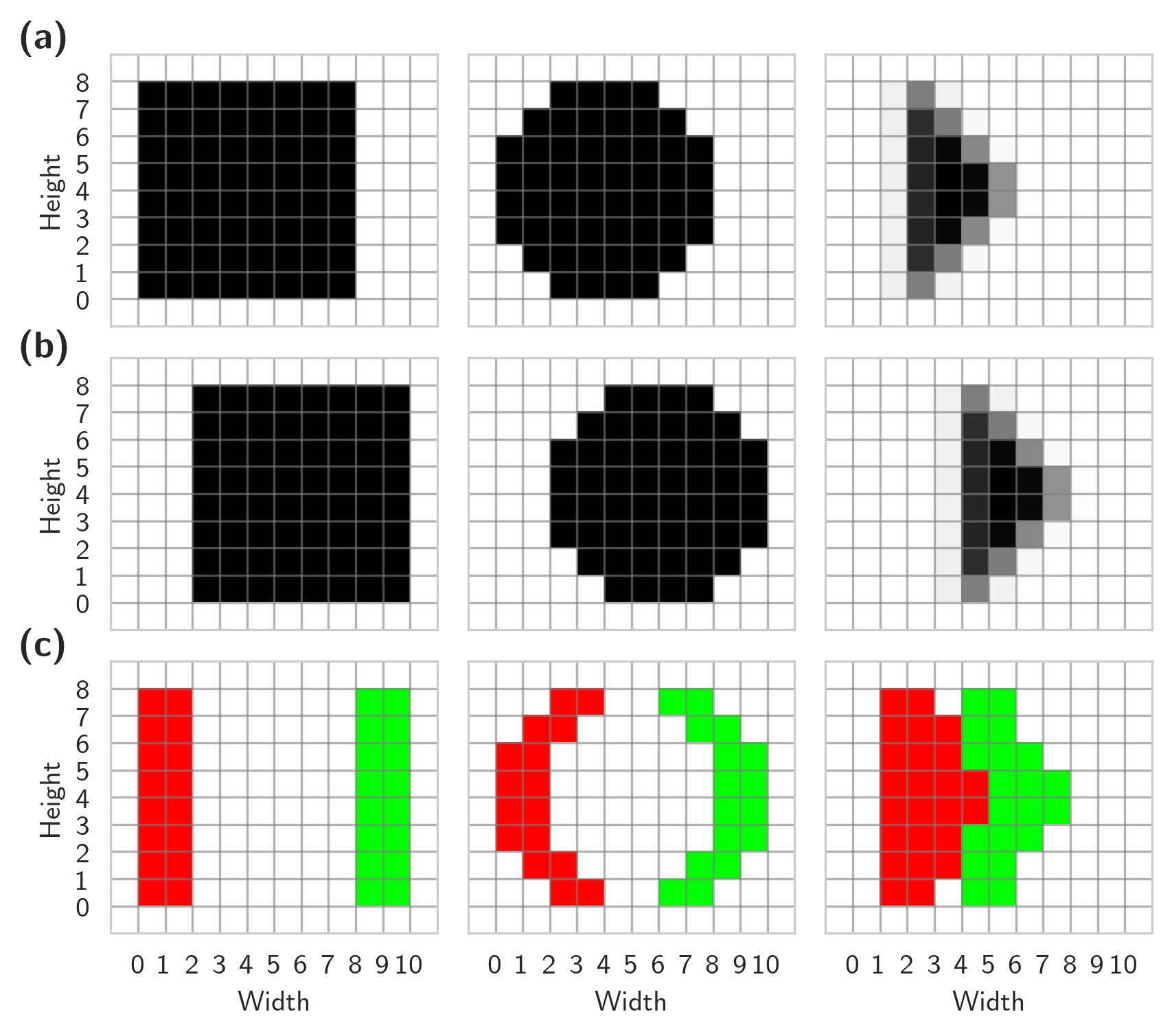}
    \caption{The three built-in shape templates in the dataset are used to generate sparse signals when moved.
    \textbf{(a)} The prototypical shapes: square, circle, and triangle.
    \textbf{(b)} The shapes translated to the right.
    \textbf{(c)} The difference between two frames generates a sparse frame with positive changes in green and negative changes in red.
    }
    \label{fig:shapes}
    \vspace{.5em}
    \centering
    \includegraphics[width=\linewidth]{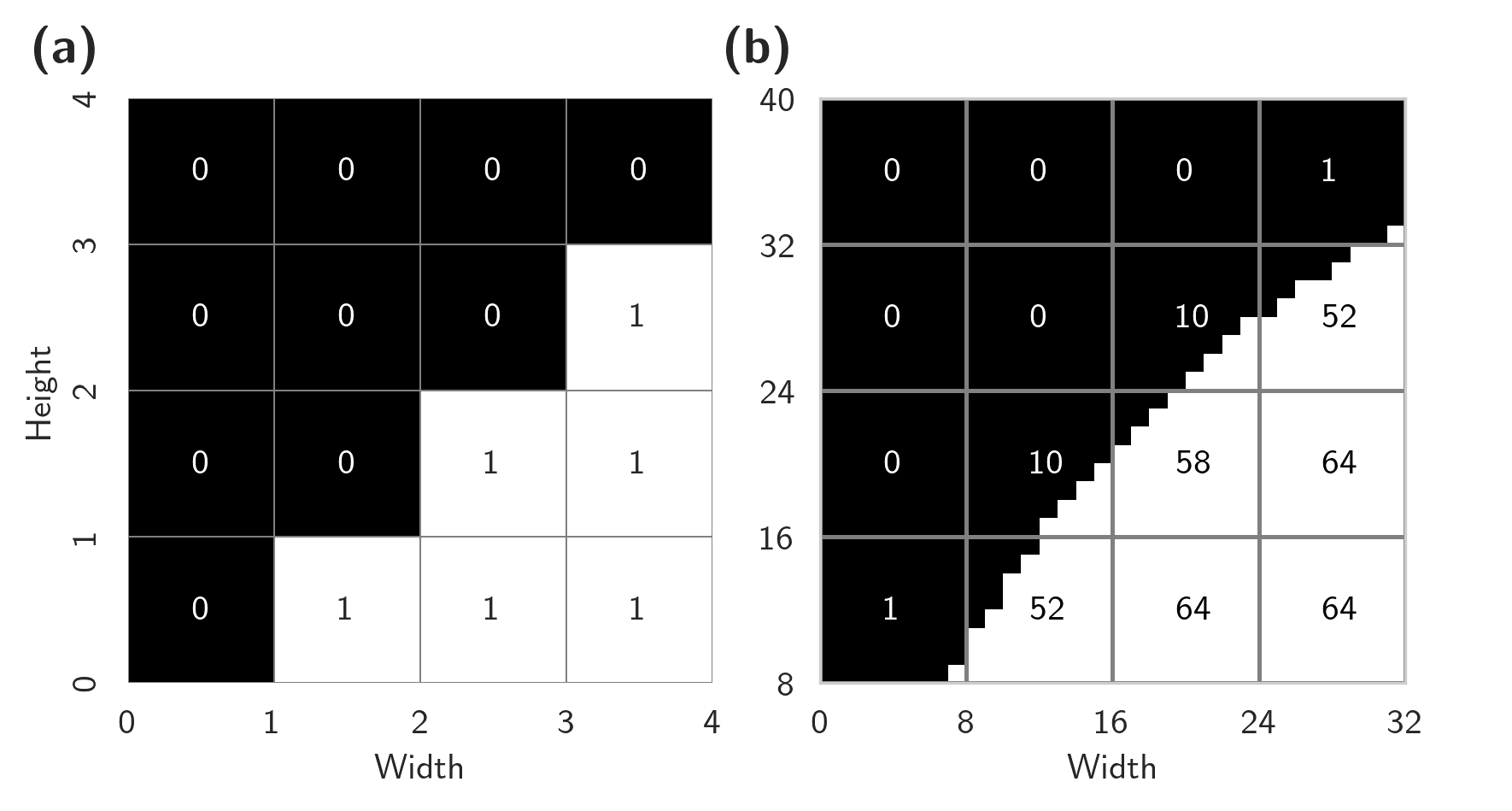}
    \caption{We upsample and integrate pixels to compensate for aliasing effects.
    \textbf{(a)} A downsampled image of the top-left part of a circle. Each pixel is either on or off.
    \textbf{(b)} When upsampling the image from \textbf{(a)} we increase the granularity of the integration.
    }
    \label{fig:upsampling}
    \vspace{-4em}
\end{wrapfigure}

Since we are operating in a discrete pixel grid, sudden displacements can cause troubling sporadic events, particularly for small resolutions.
These are known as aliasing artifacts.
The triangle in Figure \ref{fig:shapes} is a problematic example because the right-most point of the geometry is half-way between the bottom (at $0$) and the top (at $8$) in the grid, that is 4.
That point does not exist in the grid, so the shape is spatially smoothed, contrary to the square and circle that are both aligned perfectly with the grid.

To achieve sub-pixel accuracy, we operate on an upsampled version of the main pixel grid, as shown in Figure \ref{fig:upsampling}.
By using a more granular grid (Figure \ref{fig:upsampling}b), we get more granular pixel activation counts, which we relate back to the original down-sampled grid (Figure \ref{fig:upsampling}a).
Next, we integrate the activation counts $z$ by summing them across timesteps
\begin{equation} \label{eq:integrate}
    v[t] = v[t-1] + z
\end{equation}
up to a threshold $\theta_{\rm thr}$, which triggers an event in the downsampled pixel-grid
\begin{equation} \label{eq:fire}
    \text{event} = \begin{cases}
        -1 & v[t] < -\theta_{\rm thr}\\
        0 & -\theta_{\rm thr} \leqslant v[t] \leqslant \theta_{\rm thr}\\
        1 & v[t] > \theta_{\rm thr} \\
    \end{cases}
\!\!\quad \text{and resets} \!\!\quad
    v[t] = \begin{cases}
    v[t] & |v[t]| > \theta_{\rm thr} \\
    0 & {\rm else}.
    \end{cases}
\end{equation}
Note the polarization.
This is comparable to the operating principles of real event-cameras, where individual pixels pick up smaller discrete electromagnetic charges, similar to integrate-and-fire dynamics.
We add a warmup period to avoid integration artifacts and initialize the integrator uniformly by default, which gives a stochastic sampling and distributes the events more evenly over time.
\vspace{-2em}
\subsection{Defining transformations}
\begin{figure}
    \vspace{-2.5em}
    \centering
    \includegraphics[width=\linewidth]{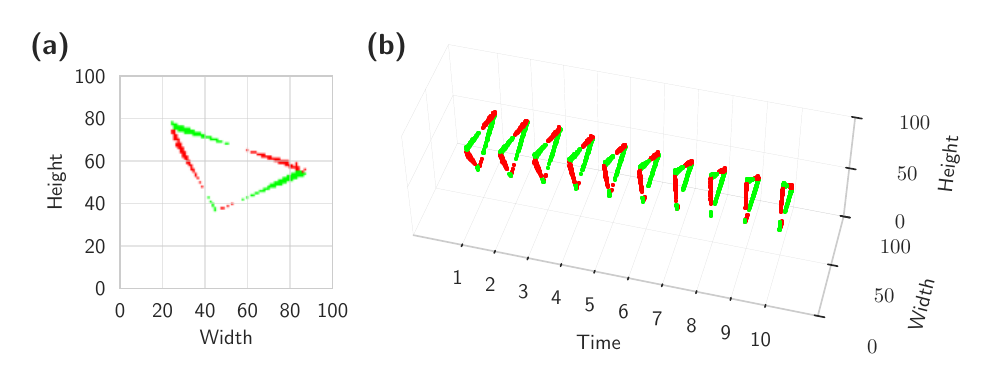}
    \vspace{-3em}
    \caption{A triangle rotated clockwise for 10 timesteps.
    \textbf{(a)} the initial frame of the movie.
    \textbf{(b)} the movie shown as a 3-dimensional point cloud.
    }
    \label{fig:3d-example}
    \vspace{-1.5em}
\end{figure}

Transforming the original signals is done with affine transformations which can be decomposed into four individual operations: translation, scaling, rotation, and shearing.
Rendered in time, this provides full control over affine transformation, Galilean transformations, and temporal scaling transformations, represented as velocity.
These properties are known to cover all possible transformations of 3-dimensional objects projected on a 2-dimensional surface, under orthographic projection assumptions \cite{Lindeberg_2023, Pedersen_Conradt_Lindeberg_2024}.
Figure \ref{fig:3d-example} visualizes the successive application of a constant rotational transformation to a triangle.
The 10 frames provide a video that isolates the effect of a constant rotation: acceleration effects can be studied if we change the transformation during the simulation.

To fully control the behavior of the simulator, the user specifies the exact behavior of the simulation, including a) the resolution, b) duration, c) upsampling behavior, d) noise configurations, and e) the initial conditions as well as velocity and acceleration for each transformation (translation, rotation, skew, and scale).
The transformational accelerations defaults to a Gaussian to produce Brownian motion, but can be overwritten with a custom function to provide arbitrary acceleration profiles.
These configurations make the simulation entirely deterministic, permitting the perfect reproduction of the data generation, and can be provided--and stored--as a textual configuration file for posterity.

\vspace{-.5em}
\subsection{Noise}
\begin{figure}
    \vspace{-1em}
    \centering
    \includegraphics[width=\linewidth]{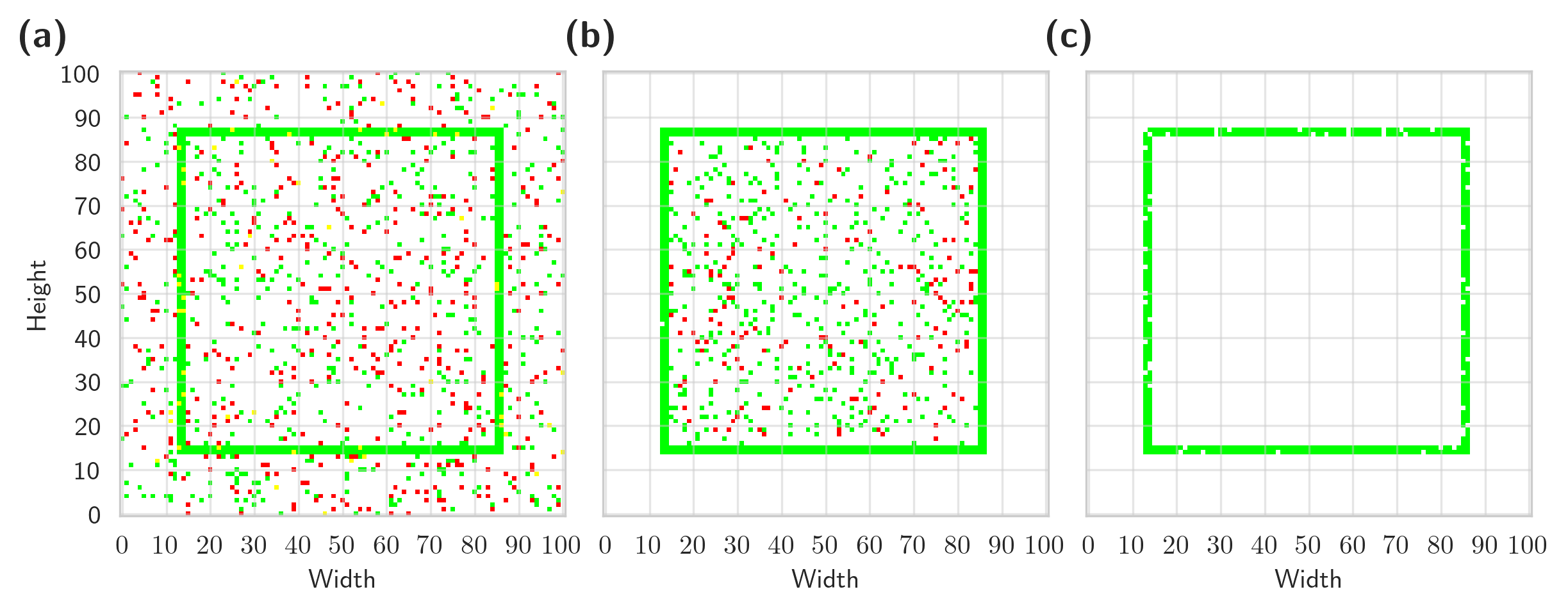}
    \vspace{-1em}
    \caption{A square scaled by two pixels and subject to three types of noises:
    \textbf{(a)} 10\% background,
    \textbf{(b)} 10\% geometry, and
    \textbf{(c)} 10\% event sampling.
    }
    \label{fig:noise-example}
    \vspace{-1em}
\end{figure}

To simulate stochasticity in the event generation process, we provide the means to control three different types of noise: background noise, shape sampling noise, and event sampling noise, illustrated in Figure \ref{fig:noise-example}.

\textit{Background noise} corresponds to noise in event-sensors where pixels randomly fire, independently of the actual stimuli.
This type of noise is useful to ensure that models generalize to pixel noise.

\textit{Shape sampling noise} appears when a pixel inside a shape does not trigger an event.
Since the events are sampled from the difference between two frames, each of which may have ``missing'' pixels, this can result in both positive and negative events occurring \textit{inside} a shape.
In real event cameras, this can occur due to material reflectivity (albedo), occlusion, or environmental lighting conditions.

\textit{Event sampling noise} determines the probability with which we sample the difference between two event frames.
This rarely occurs in real settings, but is a clean way to add noise to the event signal.

\subsection{Data Loader for model training}
We additionally provide a PyTorch Data Loader which can be used directly in the training process.
The loader picks up recordings from a directory and serves the data along with the coordinate labels for the shapes.
\vspace{-0.5em}
\section{Applications}
While the datasets generated with our method can be applied for numerous applications, GERD is primarily designed for the study of precise, geometric transformations in time and space.
Below, we highlight four example use cases and subsequently expand one of them--transformation invariance--to demonstrate the study of time scale invariance using our dataset generation method.
\vspace{-0.5em}
\subsection{Example use cases}
\begin{itemize}
\item \textbf{Mock stimulus:} When working with event-based vision pipelines, it is sometimes helpful to test the system with rudimentary stimuli before testing it with real-world data. This could be a simple triangle subject to rotation, as shown in Figure \ref{fig:3d-example}.
    \item \textbf{Transformation invariance:} When detecting objects, it is typically desirable to be invariant to certain transformations that distort the signal. By exposing the same stimulus to those transformations at varying amounts, the resulting dataset can be used to train or test invariances, as in \cite{Pedersen_Singhal_Conradt_2023}. Withholding a subset of the dataset that has been transformed by a specific amount, for instance by the largest scaling factor, provides a test for the generalization capacities for that specific transformation, as seen in \cite{Jansson_Lindeberg_2022}. An example is developed in the section below.
    \item \textbf{Transformation covariance:} Sensitivity to the magnitude of the transformation is an important property for computer vision because it optimally captures affine, Galilean, and temporal scaling operations on 2-dimensional signals \cite{Lindeberg_2023}. By controlling the velocity of the transformation, our method generates a dataset that tests for covariance properties under different transformations, as has been done in \cite{Pedersen_Conradt_Lindeberg_2024}.
    \item \textbf{Generalization to noise:} We parameterize noise in the background, geometry, and shape sampling (see Figures \ref{fig:pipeline} and \ref{fig:noise-example}). While GERD cannot simulate the exact stochastic conditions of event cameras \cite{DVS-Voltmeter}, it enables carefully controlled noise scenarios that are useful for the studying of robustness on varying signal-to-noise ratios and sparsity of events.
\end{itemize}

\subsection{Demonstration: time scale invariant network training}
\begin{figure}
    \vspace{-2em}
    \centering
    \includegraphics[width=\linewidth]{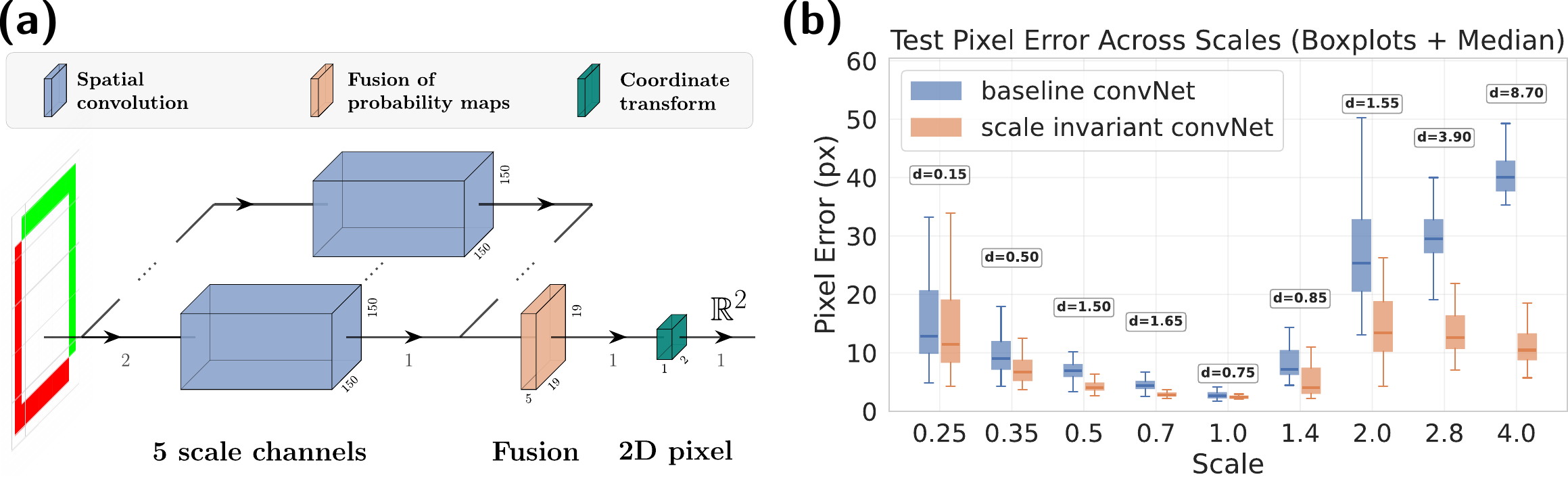}
    \caption{Network architecture and network inference quality for a scale invariant and baseline convolutional network.
    \textbf{(a)} The network receives 2 event polarity input channels, processes them through five scale channels, and fuses the scaled representations into a single coordinate prediction.
    \textbf{(b)} We train on a single scale ($s=1$) and compare both a baseline and scale invariant network across 8 unseen scales.
    The baseline network is performing well at the training scale, but generalizes worse than the scale invariant.
    Cohen's d is listed as the effect size over 400 samples per scale, where $d > 0.8$ is a large effect \cite{Cohen1969StatisticalPA}.
    }
    \label{fig:net-performance}
    \vspace{-1em}
\end{figure}

As an illustrative task, we explore the ability of a network to be invariant to scaling operations over time using a synthetically generated dataset generated with GERD.
The dataset consists of frames of a square, subject to a varying amount of translation and scaling over 60 timesteps per sample.
In the training set, these transformations are constant at a velocity of 1px, but we define test set velocities to $\{0.25, 0.35, 0.5, 0.7, 1.4, 2, 2.8, 4\}$.
For the base time scale ($s=1$), we generate 2000 datapoints.
These are divided into $70\%$ training, $15\%$ validation, and $15\%$ test sets, yielding 1400 training samples and 300 samples each for validation and testing. For the remaining time scales, each set contains 400 datapoints per time scale, used exclusively for testing.

We re-use a time scale-covariant convolutional network from \cite{Pedersen_Conradt_Lindeberg_2024} (see Figure \ref{fig:net-performance}a).
The backbone of the model is a five-layer convolutional network that processes two input channels (corresponding to polarity) through five convolutional layers with channels
\[
2 \rightarrow 12 \rightarrow 36 \rightarrow 16 \rightarrow 16 \rightarrow 1,
\]
producing a spatial heatmap that we interpret as the probability distribution for the shape coordinates.
The backbone uses leaky integrators as temporal activation functions, parameterized by a time constant $\mu$.
We duplicate the backbone into five scale channels, each with their own time constants, spanning five different temporal scales in the dataset by multiplicative factors of $2$ (see \cite{Pedersen_Conradt_Lindeberg_2024}).

The five scale channels output a two-dimensional probability distribution that is combined to a single 2D distribution using a weighted sum.
Finally, the $(x, y)$ output position is computed as the expectation (center of mass) of the fused probability distribution.
This design allows the network to combine information from multiple temporal scales, balancing fast, precise responses with slower, more stable estimates.

The model is evaluated against a baseline network with a single scale channel, which should not share the same sensitivity to temporal scaling as the model with five scale channels.

The results, shown in Figure~\ref{fig:net-performance}, demonstrate that the invariant model maintains robust and accurate localization across a wide range of transformation velocities. While performance is highest at the training velocity for both models, the invariant model substantially outperforms the baseline as the velocity deviates from this condition, highlighting its ability to generalize across temporal scales.
At lower velocities, the baseline shows comparatively strong performance, which can be attributed to the inherent low-pass filtering behavior of leaky integrators, favoring slower temporal dynamics. However, this advantage is limited, and the invariant model consistently provides better performance.

\section{Discussion}
We presented a simulation tool to generate sparse event-based recordings of carefully controlled geometries.
By carefully controlling the stimulus and the transformation they are subject to, our method permits detailed studies of geometric properties---or the lack thereof.
We presented four examples where our method has already been applied: (1) mock data for event-based vision pipelines, training data for (2) invariant and (3) covariant object detection models previously used in \cite{Pedersen_Conradt_Lindeberg_2024}, and 4) generalization to noise.

While GERD operates in a simplified 2D setting without sensor-specific effects, the interplay of spatial and temporal transformations in event-based vision remains poorly understood, and controlled synthetic data is a necessary stepping stone.
Models pre-trained or validated on GERD data can be fine-tuned on recorded event streams, using the geometric understanding acquired in simulation as an inductive bias; the ground-truth transformations further serve as a diagnostic tool for identifying which geometric properties transfer to real recordings and where the gap between simulation and reality lies.

Our work was initiated to study transformation effects systematically in event-based computer vision models, where the interplay of spatial and temporal transformations is still poorly understood.
By providing a controlled sandbox for these studies, we hope GERD can help close the performance gap between event-based and conventional frame-based vision models.

\begin{credits}
\subsubsection{\ackname}
Support from the Novo Nordisk Foundation (NNF24OC0089302) and the National Academic Infrastructure for Supercomputing in Sweden (NAISS,  2024/22-1334, 2025/5-330, and 2026/3-372).
\end{credits}

\bibliographystyle{plain}
\bibliography{bibliography}

\end{document}